\documentclass[11pt]{article}

\usepackage[final]{acl}

\usepackage{times}
\usepackage{latexsym}
\usepackage{amsmath}
\usepackage{booktabs}
\usepackage{pifont}
\usepackage{seqsplit}
\usepackage{float}
\usepackage[dvipsnames]{xcolor}

\usepackage[T1]{fontenc}

\usepackage[utf8]{inputenc}

\usepackage{microtype}

\usepackage{inconsolata}

\usepackage{graphicx}

\title{Demystifying Entropy-based Selection for \\ Chain-of-Thought Compression in Large Reasoning Models}

\author{
  Sara Candussio$^{1,*}$ 
  \quad Daniel Scalena$^{2,3,*}$ 
  \\ 
  \quad \textbf{Luca Bortolussi}$^1$ 
  \quad \textbf{Elisabetta Fersini}$^2$ 
  \quad \textbf{Malvina Nissim}$^3$ 
  \quad \textbf{Gabriele Sarti}$^{4,5}$ 
  \vspace{3mm}\\
  $^1$University of Trieste \quad $^2$University of Milano-Bicocca \vspace{1mm}\\
  $^3$CLCG, University of Groningen \quad $^4$Northeastern University 
  \quad $^5$Parallax
  \vspace{3mm}\\
  \small{$^*$Equal contribution} \vspace{1mm}\\
  \small{\texttt{sara.candussio@phd.units.it} \quad \texttt{d.scalena@rug.nl}}
}

\begin{document}
\maketitle
\begin{abstract}
Entropy-based pruning has been proposed as an effective method for compressing Chain-of-Thought (CoT) reasoning with negligible accuracy loss. We test the robustness of low- and high-entropy CoT step selection methods across various models and reasoning tasks, showing that entropy offers no advantage over random pruning in any evaluated setting. Moving from sentences to tokens, we then show that retaining low-entropy tokens seems effective only on mathematical benchmarks. We find this is due to the inherently low-entropy nature of numeric tokens, which also convey semantic content in such problems. Finally, we demonstrate that patching a subset of a few CoT tokens with their original activations recovers near-perfect full-trace performance, providing causal evidence that task information is not concentrated in a small set of CoT tokens identifiable by heuristics, but rather distributed across the full reasoning chain.\footnote{Code instructions and data at \url{https://github.com/saracandussio/Demystifying-Entropy-Selection}.}
\end{abstract}

\section{Introduction}
Inference-time scaling via Chain-of-Thought (CoT) reasoning has become the norm for improving the accuracy of large language models (LLMs) on complex multi-step tasks. However, these gains often come at the expense of unnecessarily long traces that quickly fill the models' context windows \citep{boppana2026reasoning,scalena2026commitmentboundaryprobingepiphenomenal,Qiao2025ConCISECC}, motivating the development of CoT compression methods reducing context size with limited performance loss. 

One such strategy involves treating token- or sentence-entropy as a proxy for relevance, pruning large fractions of a trace based on entropy value alone with little to no accuracy drop. Opposite claims coexist in the literature, favoring either the pruning \citep{wang20258020rulehighentropyminority,li2026making,laaouach2025haltcot,kim-etal-2026-think} or the retention \citep{huang2026pear,ton2025understanding,wang2026eat,xiong2026etrentropytrendreward} of low-entropy items.

This disagreement suggests that the entropy signal might not be as precise or indicative as presented in the literature. Given that previous works lack comparison to random baseline, one may wonder whether it is indeed entropy the reason for their reported success, or it is generic compression which is effective. Additionally, since compression methods are mostly tested on mathematical benchmarks, we do not know whether the observed success deriving from entropy-based compression strategies is a property of mathematical reasoning or generalises to non-mathematical tasks, too. 

In this work, we test whether the claimed advantage of low- and high-entropy selection for CoT compression persists under both conditions, evaluating LLMs of various sizes from three model families (\texttt{gpt-oss}, \texttt{Gemma-4}, and \texttt{Qwen3}) on mathematical, logical and commonsense reasoning tasks. We show that while at the sentence level, entropy generally provides no advantage over random selection, at the token level, an apparent advantage of low-entropy selection emerges only on mathematical benchmarks.
Our results demonstrate that this can be explained by the overlap between low-entropy tokens and content-rich number tokens, rather than by any special property of low-entropy tokens. Even for mathematical tasks, selecting number tokens is most effective only when paired with a direct patching of activations. Taken together, our results suggest that the semantic content of a reasoning trace is not concentrated in entropy-relevant locations.

\section{Related work}
\label{sec:related}

A growing body of work treats token- and sentence-level entropy as a heuristic for identifying relevant content in a CoT trace, under the premise that low-entropy tokens are predictable given the preceding context and therefore redundant, while high-entropy tokens mark points of genuine deliberation and reasoning direction. \citet{wang20258020rulehighentropyminority} distinguish between the latter, dubbed as \emph{forking tokens}, and the remaining vast majority of low-entropy tokens that merely complete reasoning already initiated at such forks. \citet{li2026making} operationalize the same intuition at sentences-level, reporting that on 50 test samples removing up to 80\% of low-entropy steps (equivalent to 45\% of tokens) leaves accuracy unchanged, while random or high-entropy removal degrades it sharply. They also argue that filtering at sentence-level is preferable to filtering at token-level, since removing individual low-entropy tokens causes a sharp performance drop that they attribute to a loss of \emph{syntactic coherence}. On the other hand, \citet{huang2026pear} show that discarding high-entropy tokens can also improve accuracy; they exploit this finding and propose a reward mechanism to bias the model to avoid excessively high-entropy tokens, leading to good compression without performance loss. 

On the role of these entropy-related tokens inside the trace, \citet{zhao2026shorthandthoughtcompressingllm} propose the distinction between \emph{structural tokens}, i.e. low-entropy tokens exhibiting low semantic content and having a prevalent syntactic role, and \emph{organic tokens}, i.e. the high-entropy ones constituting instead the main skeleton of the reasoning. This distinction is notable since low-entropy tokens are elsewhere treated as bearing the semantic content of a sentence \cite{wang20258020rulehighentropyminority,huang2026pear}. 
\citet{ton2025understanding} point out that a high-entropy step may reflect either a genuine reasoning fork or noise arising from model under-training on a given task, a failure mode that they term an \emph{unidentifiable task}. They also claim that low-entropy steps may be anchor points for subsequent reasoning, such as the end of useful computation in the CoT. \citet{wang2026eat} push this observation further: when the model is unsure of the answer, entropy exhibits high variance across different rollouts, making it an unreliable signal of the trace's convergence to the final output. They also show that trace token entropies are not informative at all, preferring the entropy at the first answer token.

Due to this instability of trace entropy as a pointwise signal, \citet{xiong2026etrentropytrendreward} note that a downward entropy trend is typically associated with longer traces and show that a trend-based reward signal outperforms a purely local, per-token entropy compression criterion. \citet{chen2026ares} observe that, although others use per-token entropy directly for threshold-based early-exit decoding \citep{laaouach2025haltcot,kim-etal-2026-think}, this signal is too noisy to threshold in isolation. As a consequence, they compute a moving average over a sliding window, treating these sustained local maxima as the critical points of a trace; reducing these points improves performance on easier problems but substantially degrades it on harder ones. 

These works offer no consensus on what entropy really captures in a reasoning trace: it is unclear whether high-entropy tokens mark genuine forks or model uncertainty, and whether low-entropy tokens carry syntax or semantics remains contested across studies. This, combined with the near-total focus on mathematical benchmarks \citep{wang20258020rulehighentropyminority,li2026making,huang2026pear,ton2025understanding,xiong2026etrentropytrendreward} and the scarcity of random baseline comparisons (present only in \citet{li2026making}), motivates the robustness assessment provided by our work. 

\section{Method}
\label{sec:method}

\paragraph{Entropy definitions}
Given a reasoning trace, we define the entropy of a token as the Shannon entropy of the model's next-token predictive distribution at that position, restricted to and renormalized over its top-$k$ log-probabilities returned at generation time. The \emph{token-level entropy} is $H_i = - \sum_{t=1}^k p_i(t)  \log p_i(t)$, where $p_i(t)$ is the normalized probability assigned to a vocabulary token $t$ at sequence position $i$. 
We compute \emph{sentence-level entropy} as the mean token-level entropy over the sentence, where sentence boundaries are identified via fixed heuristics (Section~\ref{sec:exp}).

\paragraph{Compressed CoT formatting}
The thinking region $C$ inside a reasoning trace is identified as the token span between the model's beginning- and end-of-thinking markers. A selection criterion $s$ ranks the candidate units (sentences or tokens) within $C$ and adds them in ranked order until a target token budget $b = \max(1, \min(r \cdot |C|, |C|)) $ is reached, where $r \in (0, 1]$ is the retention rate. The selected items are recomposed in the order in which they appeared in the original trace, producing a compressed CoT $\tilde{C}^s_r$.

The compressed trace is wrapped into a reconstructed input $P + \texttt{[BOT]} + \tilde{C}^s_r + \texttt{[EOT]} + S$ where $P$ is the prompt, $\texttt{[BOT]}$ and $\texttt{[EOT]}$ are model-dependent beginning- and end-of-thinking markers, $+$ denotes sequence concatenation, and $S$ is a task-specific suffix that elicits a direct answer.\footnote{When left unspecified, we use the suffix \texttt{Therefore, the answer is \textbackslash boxed\{}. Other variants are ablated in Appendix~\ref{app:suffix-ablation}, producing near-identical results.} The answer is then greedily decoded for a limited number of tokens, forcing the model to rely on a compressed reasoning trace only.

\paragraph{Selection criteria}
At sentence-level, we define the following ranking criteria: \emph{low-entropy} and \emph{high-entropy} rank sentences by their mean token entropy, respectively retaining the lowest- and the highest-entropy sentences first; \emph{numbers} ranks sentences by the fraction of tokens containing a digit,\footnote{This resembles the prompt-based \texttt{OnlyNumbers} baseline of \citet{Xia2025TokenSkipCC}, applied here in a stricter setting.} as a proxy for essential content in mathematical tasks; \emph{low-entropy no numbers} ranks by mean entropy as before, excluding numeric tokens to disentangle their effect on entropy;\footnote{As a consequence, sentences with no non-numeric tokens are excluded from the candidate pool.} \emph{random} uniformly samples whole sentences from available candidates.

At the token level, the same five criteria are recomputed at the token granularity, with two additional pattern-based selectors: \emph{newlines} and \emph{end-of-sentence} markers, retaining only token categories that coincide with sentence boundaries. This addition is motivated by the use of end-of-step token positions as probing locations for extracting information regarding the full step \citep{bogdan2026thought}.\footnote{Since numbers, newlines and end-of-sentence markers typically match few tokens per trace, we report the actual compression rate rather than the nominal one when their candidate pool is exhausted before the nominal token budget is reached.}

\paragraph{Compression protocols}
We evaluate the compressed traces obtained through different selectors $s$ under two protocols. Following prior work presented in Section~\ref{sec:related}, the compressed context $\tilde{C}^s$ is passed to the model as a new prompt to obtain the resulting answer. However, for the token-level compression analysis, we also run a forward pass over the full, uncompressed trace, caching its per-layer hidden states, and patching them into the corresponding retained positions when generating from the compressed sequence, yielding patched context $\hat{C}^s$.
Given a fixed selector $s$, $\tilde{C}^s$ and $\hat{C}^s$ have the same text, but $\tilde{C}^s$ activations are recomputed from scratch, while $\hat{C}^s$ activations match those of the selected full-CoT tokens across all layers.

\paragraph{Baselines and evaluation}
Each trace is bounded by the \emph{full CoT} performance, a theoretical upper bound on achievable accuracy. 
This is compared against the accuracy obtained under $\tilde{C}^s_r$ and $\hat{C}^s_r$, for every selector $s$ and retention rate $r$.  
We report the \emph{relative performance retention} (RPR), i.e. the ratio of a certain setting's accuracy to the \emph{full CoT accuracy} as a general measure of compression effectiveness, using exact match with ground-truth answer.\footnote{Appendix~\ref{app:semantic-verifier} additionally reports results obtained with a semantic verifier, confirming that the trends discussed below are not an artifact of the matching criterion.}
To summarize a selector's behavior across the full RPR range, we additionally report the RPR \emph{Area Under the Curve} (AUC) as a function of retention rate $r$. We use $\Delta_{\text{AUC}} = \mathrm{AUC}_s - \mathrm{AUC}_{\text{random}}$ to quantify a selector $s$'s advantage relative to the random baseline.

\section{Experimental setup}
\label{sec:exp}

\paragraph{Models and datasets} We test the generality of entropy-based compression across model families, parameter scales, and reasoning domains using six models spanning three families and multiple sizes: \texttt{gpt-oss-20b} and \texttt{gpt-oss-120b} \cite{openai2025gptoss120bgptoss20bmodel}, \texttt{gemma-4-E4B-it} and \texttt{gemma-4-26B-A4B-it} \cite{gemmateam2026gemma4technicalreport}, and \texttt{Qwen3-4B} and \texttt{Qwen3-14B} \cite{yang2025qwen3technicalreport}, all operating in reasoning mode. We test these on mathematical reasoning (AIME 2024, 2025, and 2026; \citet{aime24,aime25, aime26}; and a subset of 100, 50, and 50 questions from MATH-500; \citet{lightman2023lets}), logical puzzles (ZebraLogic; \citet{lin2025zebralogic}), and multiple-choice science questions (GPQA-Diamond; \citet{rein2024gpqa}) to ensure our findings are widely applicable across various reasoning tasks. 

\paragraph{Trace sampling}
For each prompt, 8 reasoning traces are generated with temperature $0.7$ and top-$p$ $0.9$, with a budget of 16384 max new tokens dictated by our compute availability. Traces that do not reach the end-of-thinking marker $\texttt{[EOT]}$ within this budget are discarded. Trace compression is evaluated for $r \in \{0.01, 0.05, 0.1, 0.2, 0.3, 0.4, 0.5, 0.6, 0.7, 0.8, 0.9\}$ leaving up to $64$ tokens to complete the answer after the suffix.

\paragraph{Entropy measures and sentence segmentation}
Token-level entropy $H_i$ is computed over the top-$k=20$ log-probabilities returned by the sampler at generation time.\footnote{We use $k=20$ as a proxy of the full logit distribution due to vLLM constraints, following \citet{scalena2026eager}.} Sentences are identified by splitting the thinking region at token-level sentence-ending marks: a token marks the end of a sentence if it contains a period followed by a whitespace or if it ends with a period and the following token begins with whitespace.\footnote{Exclamation and question marks are consequently not treated as sentence boundaries.}

\paragraph{Activation extraction} We use
\texttt{NNsight} \citep{fiottokaufman2025nnsightndifdemocratizingaccess} to access and extract model internals and to intervene on them. Residual stream activations are collected at each layer over the full, uncompressed trace and subsequently used to overwrite specific positions at all the layers during generation over the compressed CoT $\hat{C}^s_r$. This
approach requires no modification to model weights or architecture, and is compatible with all evaluated model families through a shared interface.

\section{Results}
\label{sec:results}
\begin{figure}[h]
\centering
\includegraphics[width=0.9\columnwidth]{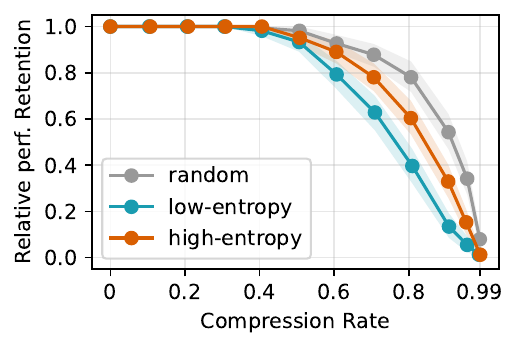}
\caption{\textbf{\emph{Random} dominates both entropy-based selectors across the entire compression range.} $\text{RPR}$ vs. Compression Rate for \texttt{gpt-oss-20b} on AIME25 for various selection strategies. Shaded bands are 95\% bootstrap Confidence Intervals (Appendix~\ref{app:bootstrap-ci}).}
\label{fig:gptoss20b-aime2025-sentence}
\vspace{-10pt}
\end{figure}

We organize our findings around five questions:

\paragraph{Does entropy-based sentence pruning outperform random selection?}
Figure~\ref{fig:gptoss20b-aime2025-sentence} reports the result of \emph{random}, \emph{low-entropy} and \emph{high-entropy} sentence-level pruning as described in Section~\ref{sec:method} on \texttt{gpt-oss-20b} on AIME25. Across the entire compression range, neither entropy-based selection criterion outperforms \emph{random}: at low compression rates all three selectors perform comparably, while at moderate-to-high compressions ($\geq 0.5$) \emph{random} retains $78$-$90\%$ of full CoT performance, compared to $40$-$79\%$ for \emph{low-entropy} and $60$-$89\%$ for \emph{high-entropy}. Entropy alone, in either direction, fails to isolate the content necessary for correct reasoning.

\begin{table}[h]
\centering
\small
\setlength{\tabcolsep}{2.5pt}
\begin{tabular}{l ccc ccc}
\toprule
& \multicolumn{3}{c}{AIME25 (math)} & \multicolumn{3}{c}{ZebraLogic (non-math)} \\
\cmidrule(lr){2-4}\cmidrule(lr){5-7}
Model & rand & $\Delta$low & $\Delta$high & rand & $\Delta$low & $\Delta$high \\
\midrule
\texttt{Qwen-4B} & \textbf{0.930} & -0.001 & -0.080 & \textbf{0.950} & -0.045 & -0.007 \\
\texttt{Qwen-14B} & 0.932 & +0.011 & -0.064 & \textbf{0.951} & -0.043 & -0.001 \\
\texttt{gemma-4B} & \textbf{0.777} & -0.001 & -0.082 & \textbf{0.936} & -0.016 & -0.018 \\
\texttt{gemma-26B} & \textbf{0.914} & -0.124 & -0.056 & \textbf{0.973} & -0.022 & -0.013 \\
\texttt{gpt-20b} & \textbf{0.872} & -0.129 & -0.060 & \textbf{0.876} & -0.026 & -0.052 \\
\texttt{gpt-120b} & \textbf{0.832} & -0.023 & -0.082 & 0.888 & +0.016 & -0.027 \\
\bottomrule
\end{tabular}
\caption{AUC of \emph{random} and $\Delta$ AUC of \emph{low} / \emph{high} relative to it over the RPR range $r\in (0,1]$, for a math-reasoning dataset (AIME25) and a non-mathematical one (ZebraLogic). \emph{random} is bolded when it outperforms both entropy-based selectors. Model names are abbreviated.}
\label{tab:auc-summary}
\vspace{-10pt}
\end{table}

\paragraph{Does this hold across model families, sizes and reasoning domains?}

Table~\ref{tab:auc-summary} reports, for two representative datasets, the AUC of the \emph{random} baseline and the delta of \emph{low-} and \emph{high-entropy} pruning relative to it, over the full RPR range. Retaining \emph{high-entropy} sentences is a poor pruning strategy regardless of model: on AIME25, $\Delta_\text{low}$ is within noise of zero for \texttt{Qwen3-14B}, \texttt{Qwen3-4B}, \texttt{gemma-4-E4B-it} and \texttt{gpt-oss-120b}, but drops sharply for \texttt{gpt-oss-20b} and \texttt{gemma-4-26B-A4B-it}. On ZebraLogic, $\Delta_\text{low}$ is negative for five of the six models, with \texttt{gpt-oss-120b} being the only exception (full breakdown in Appendix~\ref{app:sentence-full}). These results suggest that the low-entropy degradation depends on domain rather than model family, and that the entropy pruning assumption fails systematically: across different model families, scales, and domains, no selector significantly outperforms random sentence selection.

\paragraph{Is entropy the right explanation at token-level?}

\begin{figure}[t]
\centering
\includegraphics[width=\columnwidth]{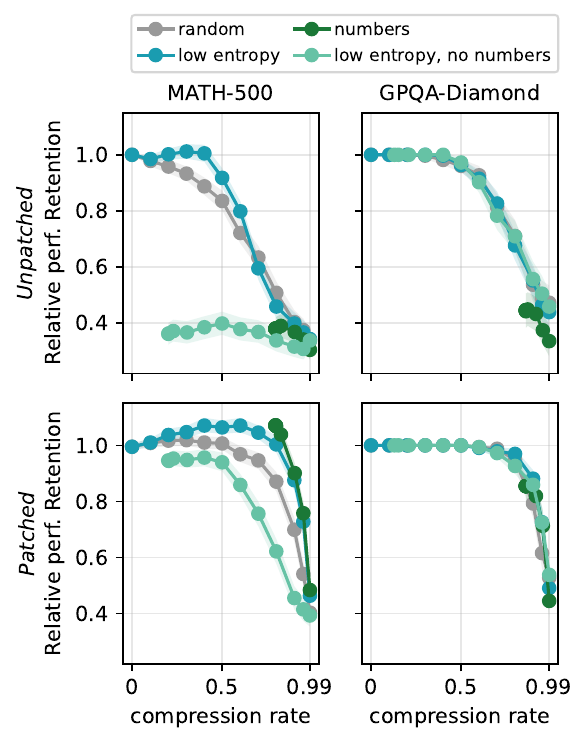}
\caption{\textbf{\emph{Low-entropy} token-level compression outperforms random selection only on mathematical tasks when patching is performed, thanks to numerical tokens.} On MATH-500, \emph{low-entropy no numbers} drops sharply below both \emph{low-entropy} and \emph{random}, while \emph{numbers} is the most performant selector even at extreme compressions. The advantage of \emph{low-entropy} on mathematical reasoning is driven by the limited choice of numeric tokens.}
\label{fig:gptoss120b-lowent-math-gpqa}
\vspace{-10pt}
\end{figure}

Unlike sentence-level pruning, token-level compression for math reasoning tasks benefits from \emph{low-entropy} selection over \emph{random} for most compression rates, while this gap vanishes on non-mathematical tasks (Figure~\ref{fig:gptoss120b-lowent-math-gpqa}, top row). We hypothesize that this asymmetry can be explained with the fact that numerical tokens exhibit significantly lower entropy in mathematical CoTs, in light of the limited vocabulary options for single-token digits.\footnote{I.e., when the context strongly implies a number, the model needs to choose only between 0-9 rather than the full set of possible tokens.} To test whether this confounds the observed trend, we distinguish between \emph{low-entropy no numbers} and \emph{numbers} settings: if numeric tokens are responsible for the gap, the latter should perform comparably to \emph{low-entropy}.

On unpatched compressed traces $\tilde{C}^s_r$, we find that \emph{low-entropy no numbers} and \emph{numbers} both perform significantly worse than low-entropy (Figure~\ref{fig:gptoss120b-lowent-math-gpqa}, top left; Table~\ref{tab:token-auc-delta} in Appendix). We attribute this difference to the lack of context fluency produced by token-level filtering, as also highlighted by \citet{li2026making}, with the resulting context often containing bare digits and arithmetic operators, or syntactic connectives stripped of their numeric content.\footnote{A concrete example is shown in Appendix~\ref{app:token-full}.} This behaviour is expected to disappear when token original activations are added to compressed forward pass: patched compression $\hat{C}^s_r$ results (Figure~\ref{fig:gptoss120b-lowent-math-gpqa}, bottom row) indeed confirm that selecting and patching only numbers in the math domain (\emph{numbers}) exceeds \emph{low-entropy} performance. Selecting instead non-numerical low-entropy tokens (\emph{low-entropy no numbers}) leads to significant degradation, even worse than \emph{random}.  
On GPQA-Diamond (Figure~\ref{fig:gptoss120b-lowent-math-gpqa}, bottom right), all strategies instead perform on par with random selection. When numerical tokens are not task-relevant content, \emph{numbers} and consequently \emph{low-entropy} selectors are comparable to \emph{random} pruning. 

\paragraph{Does token activation patching performance recover generalize to other selectors?}
Having shown that patching restores \emph{numbers}' advantage on mathematical tasks, we question whether this holds for other selectors across models and benchmarks. 

\begin{figure*}[t]
\centering
\includegraphics[width=\textwidth]{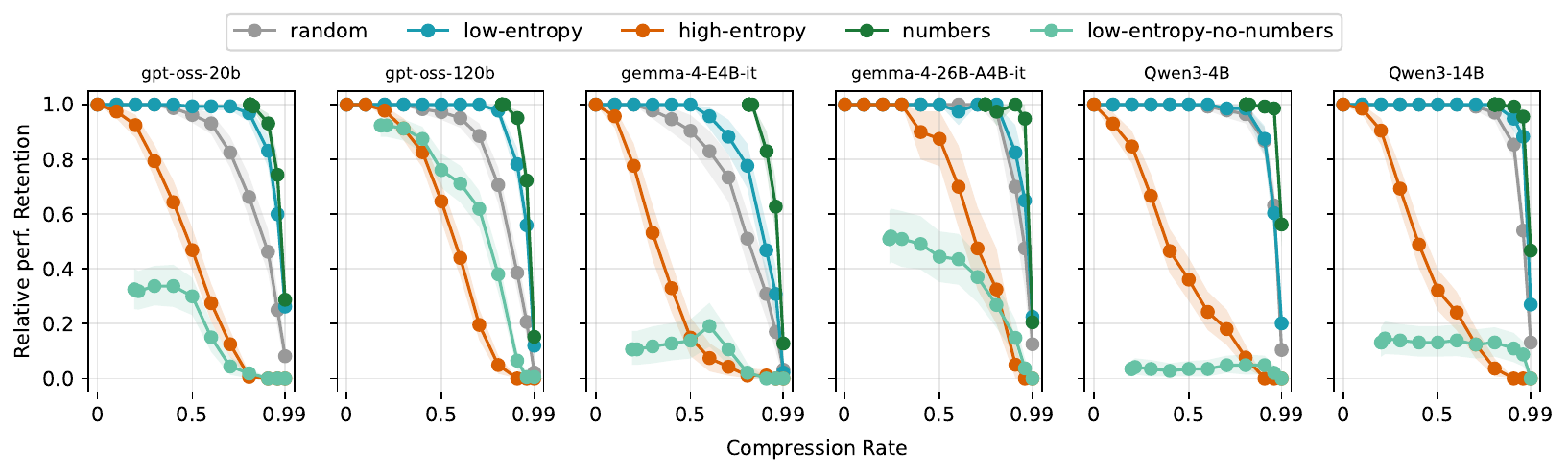}
\caption{Patching closes the gap for \emph{low-entropy} and \emph{numbers} across all models on AIME25, while \emph{high-entropy} and \emph{low-entropy no numbers} remain largely unaffected or worse.}
\label{fig:patch-strip-aime2025}
\end{figure*}

Figure~\ref{fig:patch-strip-aime2025} reports the effect of activation patching on AIME25 for all six models: \emph{low-entropy} remains close to the full trace performance up to much higher compression rates than the unpatched setting, retaining $\geq 0.9$ relative performance using $20$-$30\%$ of the original trace (compared to $50$-$70\%$ without patching). \emph{Numbers} also retains most of the trace's informativeness, almost recovering original performance using only $10$-$20\%$ of the tokens, while both \emph{low-entropy no numbers} and \emph{high-entropy} show no comparable improvement once patched. Similarly, purely pattern-based selectors (\emph{newlines} and \emph{end of sentence}, that do not carry surface-relevant content) do not perform better under patching (average $\Delta_{\text{AUC}}$ of $-0.74$ and $-0.73$ on AIME25 respectively, both worse than their unpatched versions; full results in Appendix~\ref{app:token-full-patch}). Compression with patching works only when content-rich tokens are retained: for mathematical tasks, \emph{high-entropy} tokens do not overlap with meaningful content, and \emph{low-entropy} ones owe their effectiveness to the numerical subset they include.


\paragraph{Does discarding either high- or low-entropy elements improve accuracy?}
\citet{huang2026pear} observe that removing high-entropy tokens improves original performance\footnote{Paragraph 2.3, Figure 3 of \citet{huang2026pear}: averaged across GSM8K \citep{cobbe2021gsm8k}, MATH-500, AIME24 and AMC23 \citep{wang2025stepguided} and tested on \texttt{Qwen3-4B} and \texttt{Qwen3-8B}.}, claiming that \emph{their absence reduces noise in the reasoning process}. We observe a superficially similar pattern (see left panels of Figure~\ref{fig:gptoss120b-lowent-math-gpqa} and Appendix~\ref{app:numbers-overshoot}), but attribute it to the numeric-token overlap discussed above (Section~\ref{sec:results}), since activation-patched \emph{numbers} is linked to a larger increase in accuracy. 

On the opposite, \citet{li2026making} prune low-entropy sentences\footnote{Figure 1 left of \citet{li2026making}: \texttt{DeepSeek-R1-7B} \citep{deepseekr1} on 50 samples from DeepScaleR \citep{deepscale}; \texttt{Qwen3-8B} is also used throughout the experiments.} and conclude that \emph{a vast majority of low-entropy steps are indeed redundant} and thus can be discarded without accuracy degradation, while this is not the case for high-entropy and random sentence selection. Our results do also confirm that pruning strategies can lead to improved performances (Appendix~\ref{app:numbers-overshoot}), but do also show that random selection achieves the same at comparable or even higher compression rates (Figure~\ref{fig:gptoss20b-aime2025-sentence}, Appendix~\ref{app:sentence-full}).  

The claim that entropy values (in both directions, either high- or low-) are by their own a signal for reasoning content relevance is not supported by our experiments.

\section{Conclusions}
In this work, we revisit the widespread assumption that token- or sentence-level entropy are associated to the semantic content of a reasoning trace and therefore can be used as a signal to prune a trace without accuracy drops. We test entropy-based selection against a random baseline across six models (spanning three families and two parameter scales) on both mathematical and non-mathematical reasoning tasks. 

At sentence-level, the assumption fails consistently: \emph{random} pruning dominates both \emph{low-} and \emph{high-entropy} selection across every model, scale and domain we tested, with no strategy reliably outperforming chance. At token-level, an apparent advantage for \emph{low-entropy} selection emerges on mathematical benchmarks only; we show this is not an entropy effect at all, but instead a spurious consequence of the overlap between low-entropy and numeric tokens, which alone account for the same gain. 

Selecting surface-text is however insufficient to effectively compress the trace, especially when the criterion produces an unreadable trace (e.g. consisting in bare digits or syntactic connectives only): the fresh forward pass performed by the model on this kind of context is uninformative of the reasoning process. When paired with activations, \emph{numbers} and consequently \emph{low-entropy} selectors recover near full-trace performance on mathematical benchmarks at extreme compression rates (using only $10$-$20\%$ of the original tokens), while the others show no comparable recovery. On non-mathematical datasets, no selector performs significantly better than random. 

These results suggest that traces can be compressed without performance loss even with few patched tokens if they correspond to meaningful semantic content, such as numbers in mathematical tasks. The fact that entropy-based selectors do not outperform random pruning (when purified by spurious signals such as numeric tokens on AIME and MATH-500) is an evidence that semantic content of a reasoning trace is not a property that entropy elements carry.  

\section*{Limitations}
\label{sec:limitations}
Our study is restricted to six open-weight models across three families; while this spans a discrete range of scales and training recipes, we cannot rule out that some closed, bigger (we tested up to 120B, with \texttt{gpt-oss-120b}), or simply belonging to another model family with different training strategies (e.g. from the DeepSeek series) would show a different relationship between entropy and reasoning relevant content. 
Our benchmarks, while spanning both mathematical (AIME, MATH-500) and non-mathematical (GPQA-Diamond, ZebraLogic) domains, remain focused on reasoning tasks. We also limit in number and in the size of the non-AIME subsets (50-100 questions each) and the trace length (16384 max new tokens) which may understate the variance of some reported gaps, a choice made for efficient data collection. 

The activation patching intervention requires access to model internals and full-precision forward passes over the uncompressed trace, which is not a deployable compression strategy on its own but rather a diagnostic tool used here to test whether adding contextual information to unreadable traces can help recover the full trace performance. 

Finally, our selection criteria operate at fixed retention rates that can only be determined post-hoc and that are drastically different according to the original trace length.

\bibliography{custom}

\clearpage

\appendix

\section{Suffix ablation}
\label{app:suffix-ablation}

To rule out that our choice of answer-eliciting suffix (\texttt{Therefore the answer is \textbackslash boxed \{}, footnote of Section~\ref{sec:method}) does not itself bias the reported accuracies, we run a small ablation on a subset of $N=25$ questions (we limit this test to one trace per question) drawn from one mathematical benchmark (AIME25) and one non-mathematical one (ZebraLogic) across all six models, comparing the default suffix against $K=3$ alternative phrasings: \texttt{\textbackslash boxed\{},  \texttt{Based only on the above, the best answer I can determine is \textbackslash boxed\{}, and \texttt{Given the reasoning above, in one sentence, the answer is \textbackslash boxed\{}.
We report accuracy under normal greedy decoding of the full trace, since the goal here is to rule out a suffix-induced confound on the upper bound itself rather than to re-run the full compression sweep for every suffix.

Table~\ref{tab:suffix-ablation} shows the results: on ZebraLogic, all four suffixes yield identical mean accuracy across the six models (93.3\%); on AIME25, the four suffixes vary by at most 1.4 percentage points (67.3-68.7\%), well within the cross-model standard deviation (13.1 points). This confirms that the suffix acts purely as a format cue for answer extraction rather than as a source of information that could bias the reported full-CoT accuracies.

\begin{table}[h]
\centering
\footnotesize
\begin{tabular}{p{3.8cm} cc}
\toprule
Suffix & AIME25 & ZebraLogic \\
\midrule
\texttt{Therefore, the answer is \textbackslash boxed \{} & 0.687 & 0.933 \\
\midrule
\texttt{\textbackslash boxed\{} & 0.673 & 0.933 \\
\midrule
\texttt{Based only on the above, the best answer I can determine is \textbackslash boxed\{} & 0.680 & 0.933 \\
\midrule
\texttt{Given the reasoning above, in one sentence, the answer is \textbackslash boxed\{} & 0.687 & 0.933 \\
\bottomrule
\end{tabular}
\caption{Full-CoT accuracy by suffix using greedy decoding, averaged across the six evaluated models ($N=25$ questions per model, one trace per question). Differences across suffixes are negligible on both benchmarks.}
\label{tab:suffix-ablation}
\end{table}

\section{Answer correctness assessment via a semantic verifier}
\label{app:semantic-verifier}
Throughout the main text, correctness is assessed via exact match between the extracted answer and the ground truth. 
Related works in this space adopt a range of different (and often incompatible) criteria for this same judgement, complicating any direct comparison of reported accuracies. 
\citet{wang20258020rulehighentropyminority} inherit the reward verifier of the underlying RL codebase (verl/DAPO), which applies symbolic normalization rather than raw string matching;\citet{huang2026pear} adopt the extraction and verification pipeline of \citet{yang2024qwen25mathtechnicalreportmathematical}; 
the entropy-after-\texttt{</think>} early-exiting method of \citet{wang2026eat} relies on canonical-form answer comparison via \texttt{sal.math}, the math-evaluation utilities of Hugging Face's \emph{search-and-learn} library \citep{beeching2024scalingtesttimecompute,snell2024scalingllmtesttimecompute}, itself adapted from the Qwen2.5-Math evaluation parser; \citet{chen2026ares} employ a two-stage criterion, i.e. normalized string matching on the extracted boxed answer followed by an LLM-as-judge fallback when no boxed answer is found or the normalized strings disagree.

Given this variability (and since we cover most of the models and datasets used across this body of work), we recompute all reported curves using \texttt{math-verify} (a shared semantic verifier \citep{kydlicek2024mathverify}) in addition to the exact-match criterion used in the main text, to check whether our conclusions are sensitive to the choice of matching criterion. Figures~\ref{fig:app-sentence-mathverify} and~\ref{fig:app-token-mathverify} report the sentence- and token-level full grids recomputed under this semantic criterion.

\section{Bootstrapped Confidence Intervals}
\label{app:bootstrap-ci}
Every accuracy we report in this work (a single point on a compression curve or an AUC summary) is itself an average over a noisy set of traces. To quantify measure uncertainty we use a non-parametric bootstrap: we resample the observed traces themselves and read the spread of the resulting statistic directly.

Shaded bands in Figures~\ref{fig:gptoss20b-aime2025-sentence} and \ref{fig:patch-strip-aime2025} are obtained via non-parametric bootstrap over traces, following the same procedure at every retention rate $r$ and for every (model, dataset, selector) combination. 

For a given selector $s$, let $C^s_{i, r} \in \{0, 1\}$ and $C^{\text{full}}_i \in \{0, 1\}$ denote respectively whether trace $i$ is answered correctly under the $s$-compressed CoT with retention rate $r \in (0, 1]$ and under the full CoT. The point estimate of Relative Performance Retention is
$$
\text{RPR}_r = \frac{\sum_i C^s_{i, r}}{\sum_i C^{\text{full}}_i}
$$

where the sum runs over all $n$ traces available for that (model, dataset, selector) combination, pooled across all evaluated questions. Since up to $8$ traces are sampled per question (Section~\ref{sec:exp}) and traces that do not reach $[\texttt{EOT}]$ within the token budget are discarded (Section~\ref{sec:exp}), $n$ represents the upper bound of the considered traces. We resample this set of $n$ traces with replacement $B=1000$ times; at each bootstrap iteration $b$, we draw indices $\{i_1, \dots, i_n\}$ uniformly with replacement from the original $n$ traces and recompute $\text{RPR}^{(b)}$, discarding any resample with zero denominator. The reported 95\% confidence interval is the $[2.5, 97.5]$ percentile range of the resulting bootstrapped distribution $\{\text{RPR}^{(b)}\}_{b=1}^{B}$.

For AUC and consequently $\Delta_{\text{AUC}}$ (Tables~\ref{tab:auc-summary}, \ref{tab:auc-delta-merged}), the same per-trace resampling is applied jointly across all retention rates within a given (model, dataset, selector) combination, before recomputing the trapezoidal AUC, rather than resampling independently at each retention rate. Since $n$ is fixed across $r$ for a given combination, the same resampled index set $\{i_1, \dots, i_n\}$ is reused at every retention rate within a single bootstrap iteration $b$. This matters because accuracy at neighbouring retention rates is not independent: it is computed from the same underlying traces, and a trace answered correctly at one retention rate tends to remain so at higher rates too. Resampling once per bootstrap iteration and reusing the same index set across all rates preserves this correlation, providing a more faithful interval than treating every point in the curve as an unrelated experiment.

\section{Detailed results}

\subsection{Sentence-level pruning}
\label{app:sentence-full}

Table~\ref{tab:auc-delta-merged} reports the full per-model, per-dataset breakdown of $\Delta_{\text{AUC}}$ for all four sentence-level entropy-based selectors compared to \emph{random}, both over the full retention-rate range and restricted to the aggressive-compression regime ($r \in [0.01, 0.3]$), where any gap tends to widen. Figure~\ref{fig:full-sentence} shows the corresponding curves for every model-dataset pair across all six evaluated benchmarks. Across the grid, \emph{random} is at or above every entropy-based selector in the majority of settings (bold rows in Table~\ref{tab:auc-delta-merged}); the few exceptions are discussed in Appendix~\ref{app:numbers-overshoot}. The pattern is consistent across both mathematical and non- datasets: none of the four entropy-based selectors (\emph{low}, \emph{low no numbers}, \emph{numbers}, \emph{high}) reliably outperforms random selection at the sentence level, for any model family or scale we tested.

\subsection{Token-level pruning}
\label{app:token-full}

\begin{table}[H]
\centering
\footnotesize
\textbf{Prompt:} \textit{``What is the smallest positive perfect cube that can be written as the sum of three consecutive integers?''}
\begin{tabular}{p{1.5cm} p{3.7cm} p{1cm}}
\toprule
Selector & Compressed CoT ($r=0.10$) & Answer \\
\midrule
\emph{low-} &
\scriptsize\texttt{n+1+2+2333133333\allowbreak\textasciicircum{}3\textcolor{ForestGreen}{\textbf{27}}\allowbreak\textasciicircum{}3\textasciicircum{}311\allowbreak\textasciicircum{}3\textasciicircum{}139\allowbreak\textasciicircum{}39\textasciicircum{}339\textcolor{ForestGreen}{\textbf{27}}333/3/3}
& \texttt{27} \ding{51} \\
\emph{low- no numbers} &
\scriptsize\texttt{cube as consecutive n, n++ a+ a+,\allowbreak{} a+a+ by\textasciicircum{} by by\allowbreak{} k\textasciicircum{}\textasciicircum{}\textasciicircum{}+\textasciicircum{}\textasciicircum{}\textasciicircum{}\textasciicircum{}\textasciicircum{}\textasciicircum{}\textasciicircum{}\allowbreak{} cube,+ + by by\allowbreak{} by by+x//}
& \texttt{729} \ding{55} \\
\emph{numbers} &
\scriptsize\texttt{1112123331333\allowbreak13333\textcolor{ForestGreen}{\textbf{27}}3\textcolor{ForestGreen}{\textbf{27}}33119393\allowbreak1931939312\textcolor{ForestGreen}{\textbf{7}}\textcolor{ForestGreen}{\textbf{27}}3\textcolor{ForestGreen}{\textbf{27}}}
& \texttt{729} \ding{55} \\
\bottomrule
\end{tabular}
\caption{Tokens retained by each selector on a short MATH-500 excerpt at retention rate $r=0.10$, and the model's greedy-decoded (unpatched) answer from each compressed text. \textcolor{ForestGreen}{\textbf{Green}} marks literal occurrences of the correct answer's digits (\texttt{27}) within the reconstructed text. \emph{low-entropy} generates the correct answer; \emph{low-entropy no numbers} and \emph{numbers} both converge on the same incorrect answer, $27^3=729$, despite \emph{numbers} also containing the substring \texttt{27} multiple times, showing that lexical presence of the correct digits is not, by itself, sufficient without the surrounding context.}
\label{tab:selector-example}
\end{table}

Similarly to the previous section, Table~\ref{tab:token-auc-delta} reports $\Delta_{\text{AUC}}$ for every token-level selector relative to \emph{random}, across all models and datasets (patched results are reported separately in the next subsection). Figure~\ref{fig:app-token-grid} shows the corresponding full grid.

Focusing on the three AIME benchmarks (2024, 2025, 2026) only, the \emph{low-entropy} advantage over random is consistent for four of the six models (\texttt{Qwen3-14B}, \texttt{Qwen3-4B}, \texttt{gemma-4-E4B-it}, \texttt{gemma-4-26B-A4B-it}), while both \texttt{gpt-oss} model scales show negligible or reversed effect. We attribute this recurring gap to the overlap between low-entropy and numeric tokens rather than to entropy itself, as argued in Section~\ref{sec:exp} and further isolated in the worked example below.

To make concrete how these three selectors diverge in practice, Table~\ref{tab:selector-example} shows the tokens retained by each on an example of a \texttt{gpt-oss-20b}'s MATH-500 trace at a fixed retention rate ($r=0.1$). \emph{low-entropy} retains a mix of numeric tokens and low-information connectives; once numeric tokens are excluded, \emph{low-entropy no numbers} is left choosing almost exclusively syntactic filler, degrading the surface trace to a sequence of connectives stripped of computational content; \emph{numbers} instead retains only digits and operators, producing a terse but computationally dense fragment. This illustrates why \emph{low-entropy no numbers} tracks \emph{numbers} so poorly once the numeric overlap is removed (Section~\ref{sec:exp}): what remains is syntax, not content.


\subsection{Token-level pruning with activation patching}
\label{app:token-full-patch}

\begin{figure*}[h]
\centering
\includegraphics[width=0.85\textwidth]{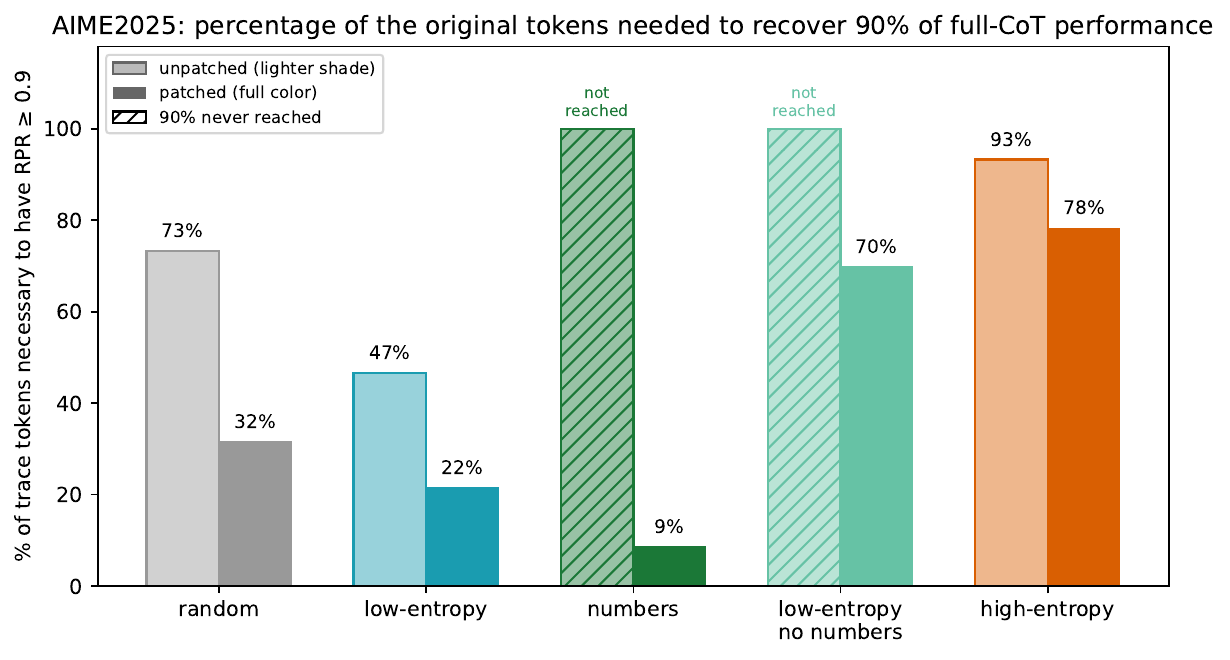}
\caption{Minimum percentage of trace tokens needed to reach and sustain a $\text{RPR}$ of at least $0.9$ on AIME25, averaged across all six models, comparing patched (full colour) and unpatched (lighter colour). Hatched bars (capped at 100\%) indicate that an unpatched compression using that selector is unable to retain the full trace performance. Patching improves every selector to some degree (including \emph{random}, from 73\% to 32\%), but the effect is markedly uneven: \emph{numbers} moves from never reaching the threshold unpatched to the most efficient recovery of all once patched (9\%), while \emph{low-entropy no numbers} only partially recovers (70\%) and \emph{high-entropy} barely improves (93\% to 78\%).}
\label{fig:patch-retention-threshold}
\end{figure*}

Figure~\ref{fig:patch-retention-threshold} makes the picture from Table~\ref{tab:token-auc-delta-patch} concrete in terms of retention budget, averaging the minimum retention rate needed to reach $\text{RPR} \geq 0.9$ across all six models on AIME25. Patching improves every selector to some degree, including \emph{random}, whose own retention requirement drops from 73\% to 32\% of the trace once patched. This is expected: patching restores full-context activations at every retained position regardless of which positions were chosen, so it should help any selector to some extent, including one that discards content without regard to relevance. The relevant question is therefore not whether patching helps, but how much, and for which selectors it changes the qualitative picture rather than merely shifting it.

\emph{numbers} and \emph{low-entropy no numbers} never reach $\text{RPR} \geq 0.9$ at any tested retention rate without patching, on average (their unpatched bars are capped at 100\% in the figure, meaning that the full trace is required to close the gap). Patching changes this qualitatively: \emph{numbers} recovers the threshold using only 9\% of the trace, the most efficient recovery among all tested selectors, while \emph{low-entropy no numbers} still requires 70\%, only a partial recovery. \emph{low-entropy}, whose unpatched advantage is itself largely inherited from its overlap with numeric tokens (Section~\ref{sec:exp}), needs 47\% of the trace unpatched and 22\% patched, a proportional improvement comparable to \emph{random}'s, rather than the qualitative jump seen for \emph{numbers}. \emph{high-entropy} weakly improves (93\% to 78\%), remaining the least efficient selector in both conditions. 

\section{\emph{Numbers} selection exceeds full trace performance}
\label{app:numbers-overshoot}
Section~\ref{sec:exp} notes that there exists some token-level cases in which the compressed CoT performance exceeds full-CoT ones ($\text{RPR}>1$). 

\begin{figure*}[h]
\centering
\includegraphics[width=\textwidth]{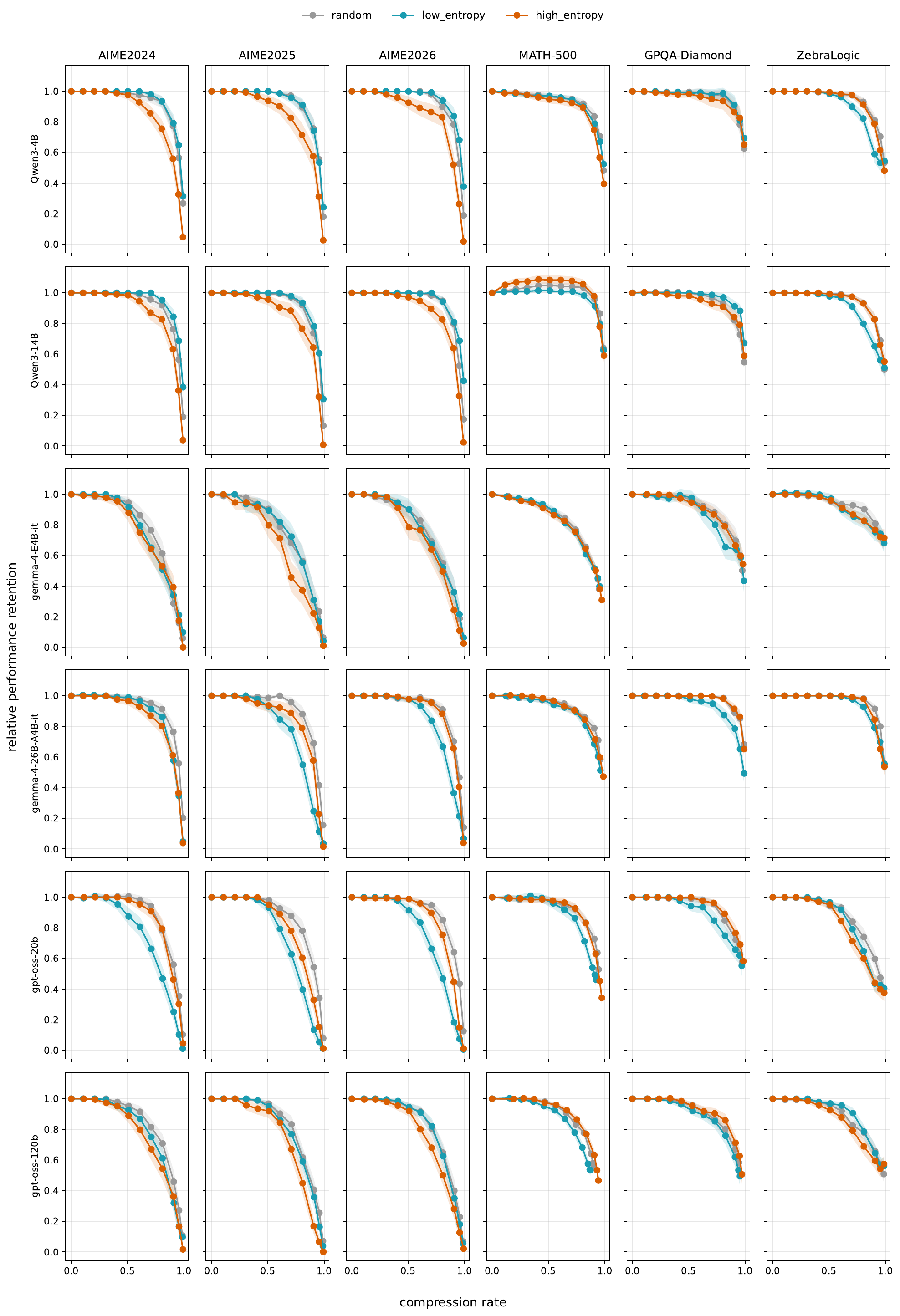}
\caption{\textbf{Sentence-level} full grid: Relative Performance Retention vs.\ compression rate for all six models (rows) $\times$ six datasets (columns), selectors \emph{random}, \emph{low-entropy}, \emph{high-entropy}. Shaded bands are 95\% bootstrap CIs (Appendix~\ref{app:bootstrap-ci}). \emph{Random} is at or above both entropy-based selectors across nearly every panel, consistent with Table~\ref{tab:auc-delta-merged}.}
\label{fig:full-sentence}
\end{figure*}

\begin{figure*}[t]
\centering
\includegraphics[width=\textwidth]{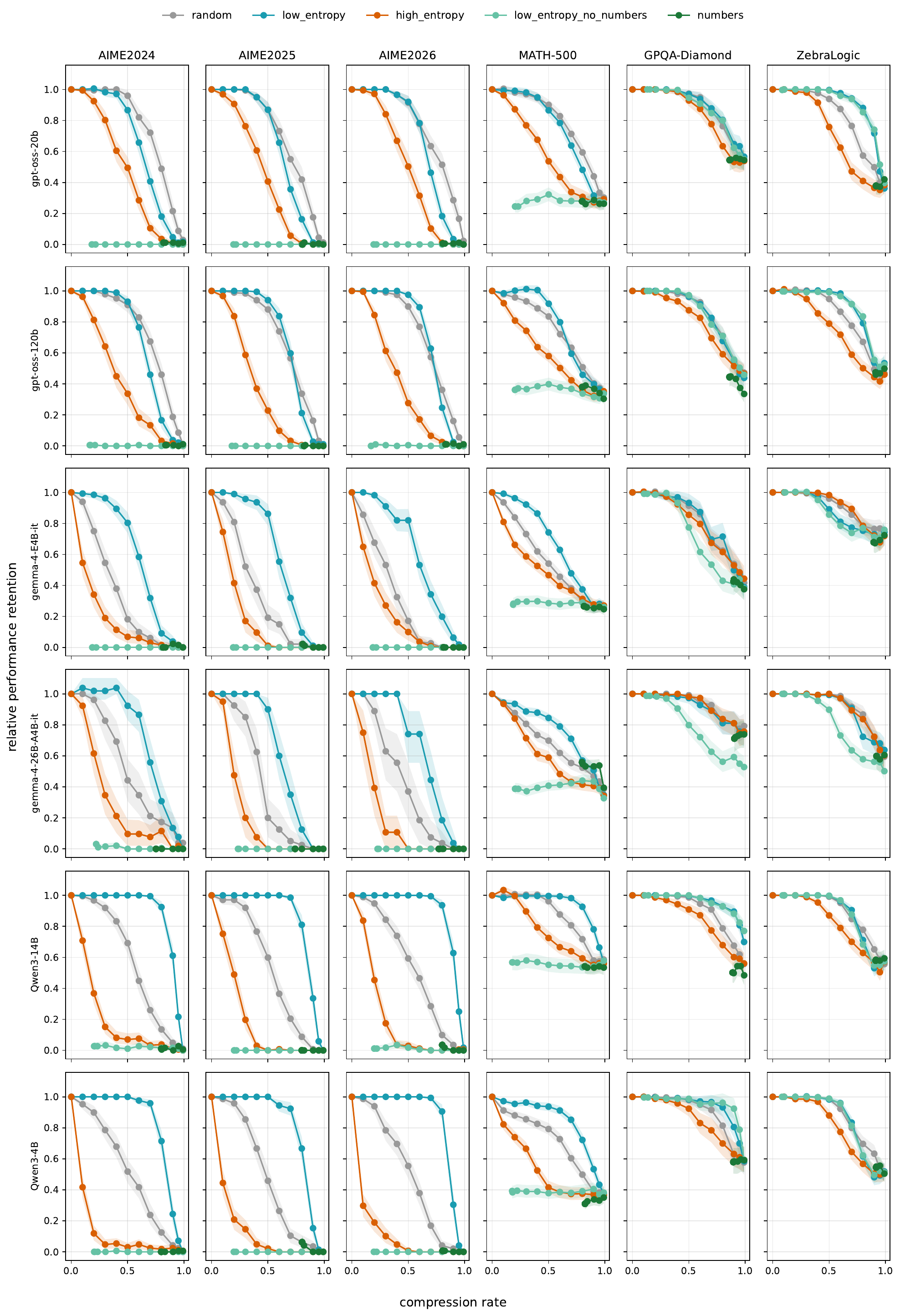}
\caption{\textbf{Token-level full grid, no patching:} Relative Performance Retention vs.\ compression rate, selectors \emph{random}, \emph{low-entropy}, \emph{high-entropy}, \emph{low-entropy no numbers}, \emph{numbers}. Model ordering: \texttt{gpt-oss-20b, gpt-oss-120b, gemma-4-E4B-it, gemma-4-26B-A4B-it, Qwen3-14B, Qwen3-4B}.}
\label{fig:app-token-grid}
\end{figure*}

\begin{figure*}[t]
\centering
\includegraphics[width=\textwidth]{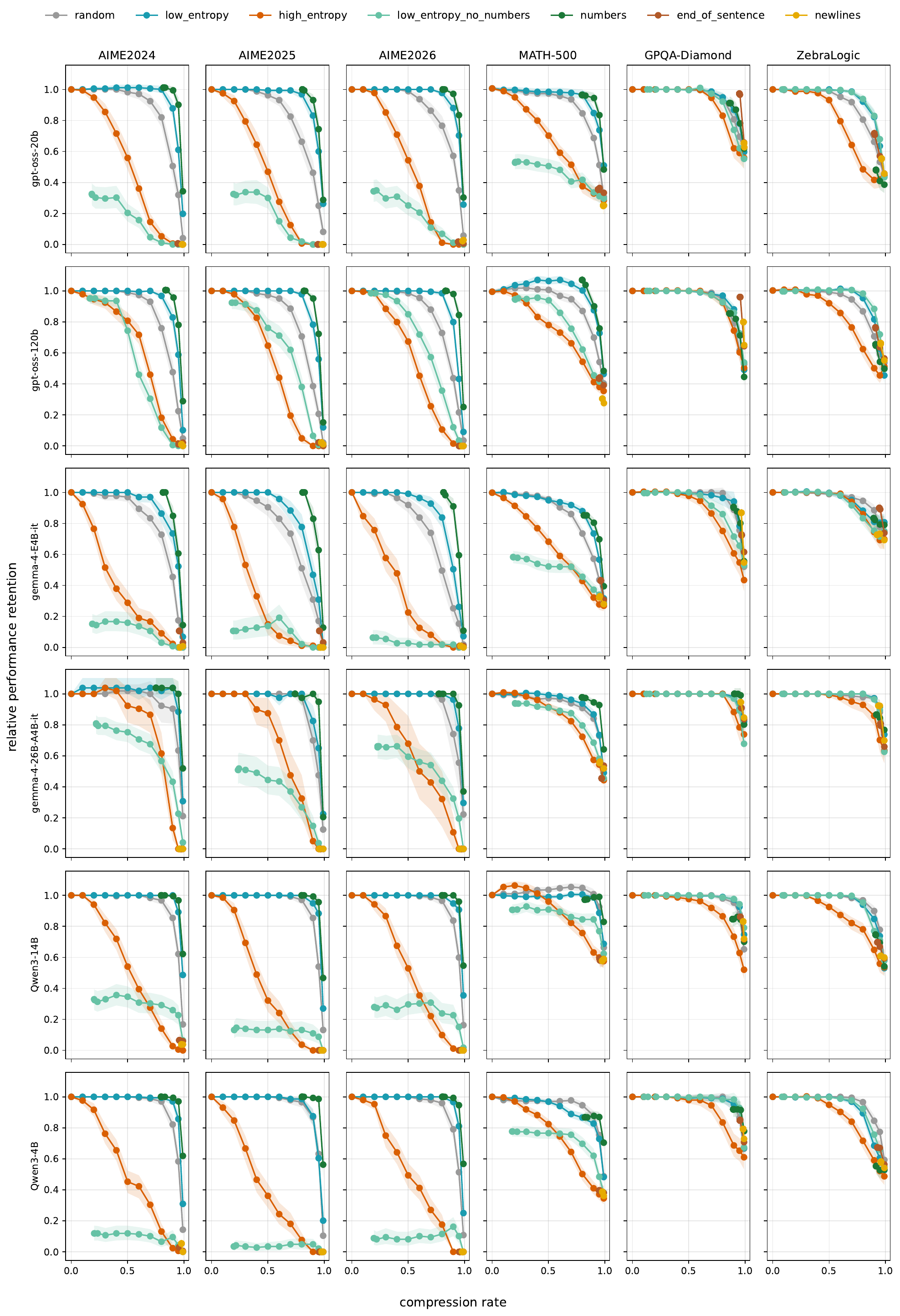}
\caption{\textbf{Token-level patched full grid:} Relative Performance Retention vs.\ compression rate, selectors \emph{random}, \emph{low-entropy}, \emph{high-entropy}, \emph{low-entropy no numbers}, \emph{numbers}. Model ordering: \texttt{gpt-oss-20b, gpt-oss-120b, gemma-4-E4B-it, gemma-4-26B-A4B-it, Qwen3-14B, Qwen3-4B}.}
\label{fig:app-token-grid-patched}
\end{figure*}

\begin{figure*}[h]
\centering
\includegraphics[width=\textwidth]{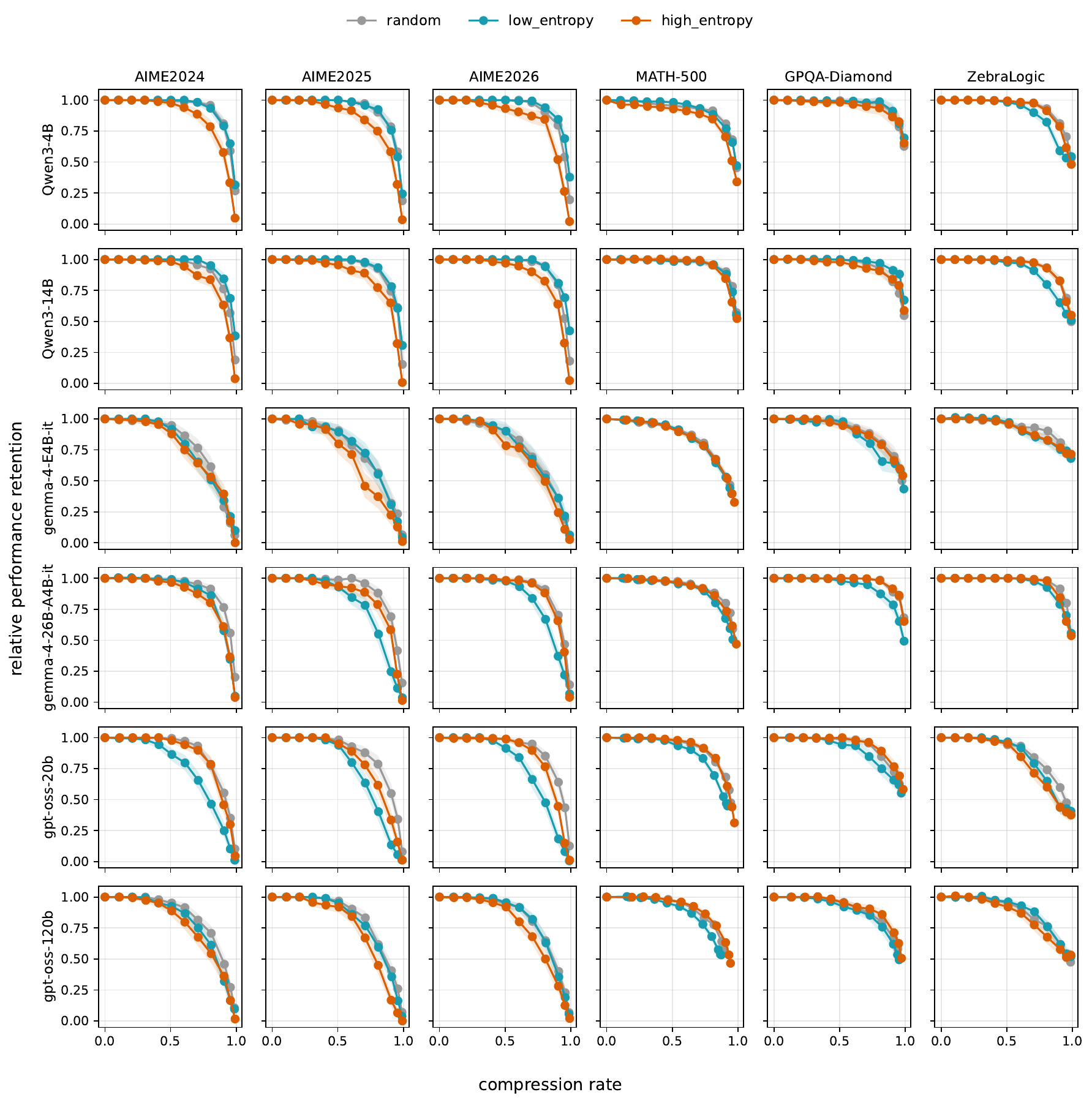}
\caption{Sentence-level full grid under \texttt{math-verify}: Relative Performance Retention vs.\ compression rate for all six models (rows) $\times$ six datasets (columns), selectors \emph{random}, \emph{low-entropy}, \emph{high-entropy}. The pattern matches Figure~\ref{fig:full-sentence} (exact-match): \emph{random} is at or above both entropy-based selectors in the large majority of panels, confirming that the sentence-level results are not an artifact of the matching criterion.}
\label{fig:app-sentence-mathverify}
\end{figure*}

\begin{figure*}[h]
\centering
\includegraphics[width=\textwidth]{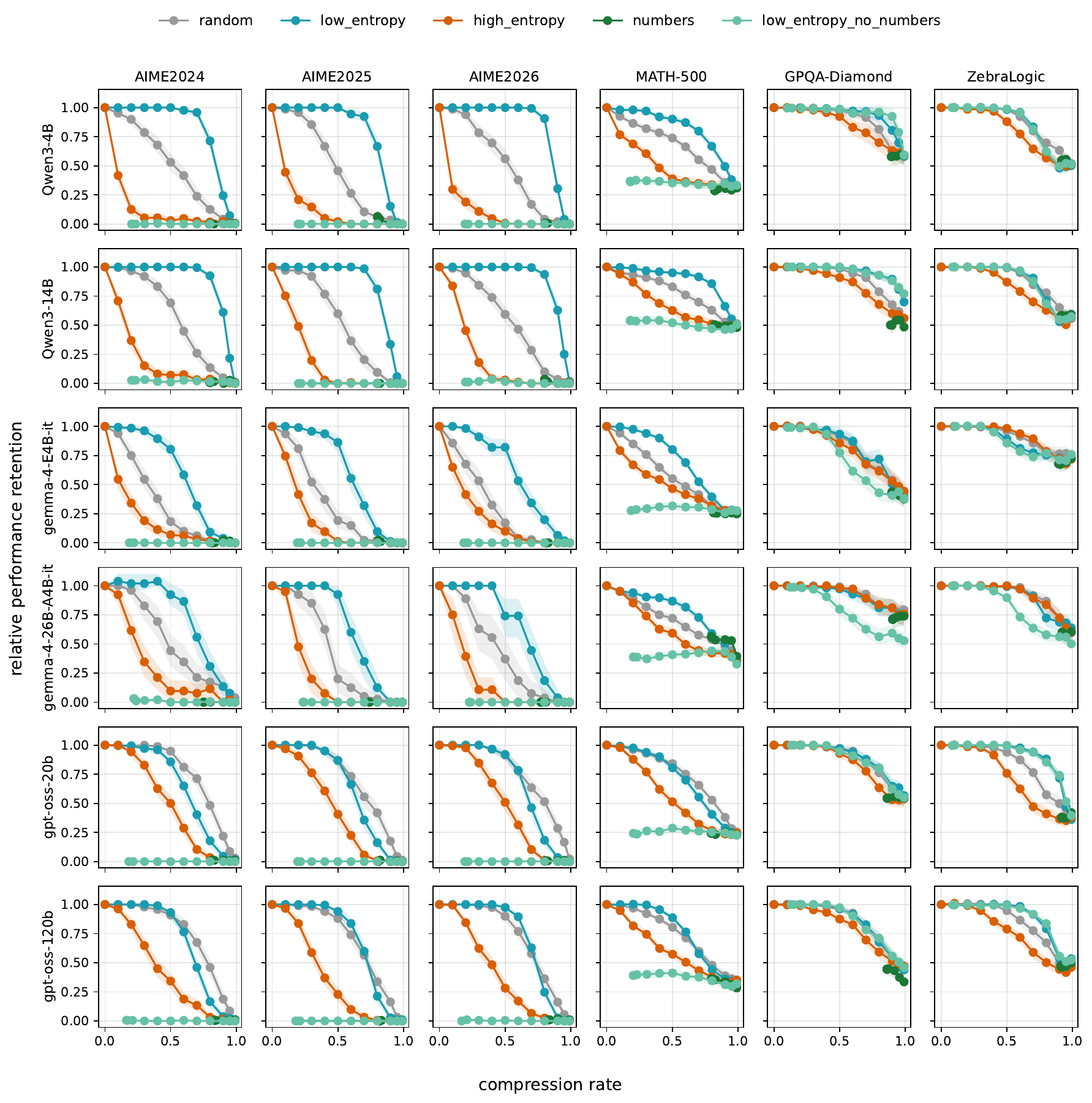}
\caption{Token-level full grid under \texttt{math-verify} (no patching): Relative Performance Retention vs.\ compression rate for all six models (rows) $\times$ six datasets (columns), selectors \emph{random}, \emph{low-entropy}, \emph{high-entropy}. The pattern matches Figure~\ref{fig:app-token-grid} (exact-match): \emph{low-entropy} exceeds \emph{random} on mathematical benchmarks only.}
\label{fig:app-token-mathverify}
\end{figure*}

Table~\ref{tab:overshoot} reports, separately for token-level selection split by patching condition, the share of all tested (model, dataset, compression rate) points for which the point estimate exceeds 1.0, and, among only those overshooting points, the share for which the 95\% CI lies entirely above 1.0. These curves are visible directly in Figures~\ref{fig:full-sentence} (sentence-level), \ref{fig:app-token-grid} (token-level), and \ref{fig:app-token-grid-patched} (token-level, patched), where panels occasionally rise slightly above the $\text{RPR}=1$ line.

\begin{table}[h]
\centering
\small
\begin{tabular}{lccc}
\toprule
Selector & \% $>1$ & \% sig.\ of these & mean \\
\midrule
\multicolumn{4}{l}{\textit{sentence-level}} \\
\midrule
high- & \textbf{2.8\%} & \textbf{66.7\%} & 1.050 \\
low- & 3.5\% & 20.0\% & 1.008 \\
low- no-num & 3.5\% & 26.7\% & 1.008 \\
num & \textbf{4.6\%} & 25.0\% & 1.015 \\
\midrule
\multicolumn{4}{l}{\textit{token-level, no patch}} \\
\midrule
high- & 0.7\% & 33.3\% & 1.016 \\
low- & \textbf{2.5\%} & 0.0\% & 1.013 \\
low- no-num & 0.5\% & 0.0\% & 1.003 \\
num & 0.0\% & --- & --- \\
\textbackslash n & 0.0\% & --- & --- \\
eos & 0.0\% & --- & --- \\
\midrule
\multicolumn{4}{l}{\textit{token-level, patched}} \\
\midrule
high- & 4.4\% & 15.8\% & 1.016 \\
low- & \textbf{9.3\%} & \textbf{17.5\%} & 1.020 \\
low- no-num & 0.7\% & 0.0\% & 1.005 \\
num & \textbf{8.6\%} & \textbf{24.3\%} & 1.031 \\
\textbackslash n & 0.0\% & --- & --- \\
eos & 0.0\% & --- & --- \\
\bottomrule
\end{tabular}
\caption{Overshoot summary, per level, selector, and (for token-level) patching condition, aggregated over all models and datasets. \% $>1$: share of all tested points whose point estimate exceeds full-CoT accuracy. \% sig.\ of these: among only the points exceeding 1.0, the share whose 95\% bootstrap CI lies entirely above 1.0. \emph{newlines} (\textbackslash n) and \emph{end of sentence} (eos) never overshoot under either condition.}
\label{tab:overshoot}
\end{table}

The only selector that is both frequent and reliably above full-trace performance is \emph{numbers} under activation patching: numeric-token activations occasionally carry enough task-relevant information on their own to match or marginally exceed full-trace performance, but only once patching supplies the context those activations would otherwise lack. No selector improves original performance \emph{without} patching and across models; the single isolated case of sentence-level \emph{high-entropy} is limited to \texttt{Qwen3-14B} on MATH-500 and therefore cannot be used to argue that either entropy extreme uniquely identifies uninformative content.


\begin{table*}[h]
\centering
\small
\setlength{\tabcolsep}{2.5pt}
\scalebox{0.9}{
\begin{tabular}{ll ccccc ccccc}
\toprule
& & \multicolumn{5}{c}{Full range $r\in[0.01,1.0]$} & \multicolumn{5}{c}{Aggressive compression $r\in[0.01,0.3]$} \\
\cmidrule(lr){3-7}\cmidrule(lr){8-12}
Model & Dataset & random & $\Delta$low & $\Delta$low\textsubscript{numbers} & $\Delta$numbers & $\Delta$high & random & $\Delta$low & $\Delta$low\textsubscript{numbers} & $\Delta$numbers & $\Delta$high \\
\midrule
Qwen3-4B & AIME24 & 0.93 & +0.01 & +0.01 & -0.04 & -0.07 & 0.79 & +0.02 & +0.02 & -0.14 & -0.19 \\
 & AIME25 & \textbf{0.93} & -0.00 & -0.00 & -0.01 & -0.08 & \textbf{0.77} & -0.00 & -0.00 & -0.02 & -0.18 \\
 & AIME26 & 0.93 & +0.02 & +0.01 & +0.01 & -0.08 & 0.78 & +0.07 & +0.05 & +0.03 & -0.16 \\
 & MATH-500 & \textbf{0.94} & -0.01 & -0.01 & -0.03 & -0.03 & \textbf{0.84} & -0.02 & -0.02 & -0.07 & -0.06 \\
 & GPQA-D & \textbf{0.97} & +0.00 & +0.00 & -0.01 & -0.01 & 0.90 & +0.02 & +0.01 & -0.04 & -0.02 \\
 & Zebra & \textbf{0.95} & -0.05 & -0.05 & -0.11 & -0.01 & \textbf{0.84} & -0.13 & -0.14 & -0.25 & -0.03 \\
\midrule
Qwen3-14B & AIME24 & 0.93 & +0.02 & +0.02 & -0.03 & -0.05 & 0.78 & +0.07 & +0.07 & -0.10 & -0.12 \\
 & AIME25 & 0.93 & +0.01 & +0.01 & -0.01 & -0.06 & 0.78 & +0.03 & +0.02 & -0.03 & -0.15 \\
 & AIME26 & 0.94 & +0.01 & +0.01 & +0.01 & -0.06 & 0.80 & +0.05 & +0.03 & +0.03 & -0.14 \\
 & MATH-500 & 1.01 & -0.03 & -0.03 & -0.03 & +0.03 & \textbf{0.96} & -0.05 & -0.04 & -0.09 & +0.00 \\
 & GPQA-D & 0.95 & +0.02 & +0.02 & -0.01 & -0.01 & 0.85 & +0.07 & +0.08 & +0.01 & +0.00 \\
 & Zebra & \textbf{0.95} & -0.04 & -0.04 & -0.10 & -0.00 & \textbf{0.85} & -0.12 & -0.12 & -0.24 & -0.00 \\
\midrule
gemma-4-E4B-it & AIME24 & 0.80 & -0.02 & -0.03 & +0.05 & -0.03 & 0.45 & -0.03 & -0.04 & +0.13 & -0.02 \\
 & AIME25 & 0.78 & -0.00 & -0.02 & +0.10 & -0.08 & 0.43 & -0.01 & -0.03 & +0.21 & -0.15 \\
 & AIME26 & 0.78 & -0.00 & -0.03 & +0.16 & -0.04 & 0.43 & -0.01 & -0.03 & +0.36 & -0.07 \\
 & MATH-500 & 0.82 & -0.01 & -0.01 & +0.01 & -0.01 & 0.59 & -0.02 & -0.03 & +0.02 & -0.02 \\
 & GPQA-D & 0.90 & -0.03 & -0.04 & +0.02 & -0.01 & 0.74 & -0.09 & -0.09 & +0.04 & -0.01 \\
 & Zebra & \textbf{0.94} & -0.02 & -0.02 & -0.06 & -0.02 & \textbf{0.84} & -0.05 & -0.06 & -0.11 & -0.05 \\
\midrule
gemma-4-26B-A4B-it & AIME24 & 0.93 & -0.04 & -0.04 & +0.01 & -0.05 & 0.78 & -0.12 & -0.13 & +0.01 & -0.13 \\
 & AIME25 & \textbf{0.91} & -0.12 & -0.14 & +0.00 & -0.06 & 0.72 & -0.31 & -0.35 & +0.03 & -0.11 \\
 & AIME26 & 0.92 & -0.08 & -0.09 & +0.04 & -0.01 & 0.74 & -0.24 & -0.26 & +0.11 & -0.04 \\
 & MATH-500 & \textbf{0.92} & -0.03 & -0.02 & +0.00 & -0.01 & \textbf{0.81} & -0.07 & -0.06 & -0.00 & -0.05 \\
 & GPQA-D & 0.98 & -0.04 & -0.04 & +0.01 & +0.00 & 0.92 & -0.12 & -0.11 & +0.02 & +0.01 \\
 & Zebra & \textbf{0.97} & -0.02 & -0.02 & -0.03 & -0.01 & \textbf{0.91} & -0.07 & -0.07 & -0.10 & -0.04 \\
\midrule
gpt-oss-20b & AIME24 & \textbf{0.89} & -0.13 & -0.14 & -0.08 & -0.02 & \textbf{0.64} & -0.28 & -0.30 & -0.18 & -0.04 \\
 & AIME25 & \textbf{0.87} & -0.13 & -0.14 & -0.05 & -0.06 & \textbf{0.62} & -0.33 & -0.34 & -0.12 & -0.17 \\
 & AIME26 & \textbf{0.90} & -0.14 & -0.17 & -0.07 & -0.05 & \textbf{0.70} & -0.36 & -0.40 & -0.17 & -0.14 \\
 & MATH-500 & \textbf{0.92} & -0.04 & -0.05 & -0.02 & -0.02 & \textbf{0.77} & -0.13 & -0.14 & -0.04 & -0.07 \\
 & GPQA-D & 0.93 & -0.04 & -0.03 & -0.03 & +0.01 & 0.78 & -0.07 & -0.05 & -0.03 & +0.04 \\
 & Zebra & \textbf{0.88} & -0.03 & -0.02 & -0.09 & -0.05 & \textbf{0.66} & -0.09 & -0.08 & -0.19 & -0.12 \\
\midrule
gpt-oss-120b & AIME24 & \textbf{0.84} & -0.04 & -0.05 & -0.04 & -0.07 & \textbf{0.55} & -0.10 & -0.12 & -0.09 & -0.13 \\
 & AIME25 & \textbf{0.83} & -0.02 & -0.03 & -0.01 & -0.08 & \textbf{0.51} & -0.05 & -0.06 & -0.02 & -0.18 \\
 & AIME26 & 0.83 & -0.01 & -0.02 & +0.02 & -0.06 & 0.51 & -0.03 & -0.04 & +0.05 & -0.12 \\
 & MATH-500 & \textbf{0.89} & -0.03 & -0.03 & -0.02 & +0.00 & \textbf{0.70} & -0.07 & -0.06 & -0.03 & -0.01 \\
 & GPQA-D & 0.90 & -0.02 & -0.01 & -0.04 & +0.02 & 0.72 & -0.04 & -0.02 & -0.03 & +0.04 \\
 & Zebra & 0.89 & +0.02 & +0.03 & -0.03 & -0.03 & 0.70 & +0.02 & +0.03 & -0.09 & -0.05 \\
\bottomrule
\end{tabular}
}
\caption{AUC of \emph{random} and delta of each entropy-based selector $s$ relative to it ($\Delta = \mathrm{AUC}_s - \mathrm{AUC}_{\text{random}}$), sentence-level selection, shown for the full retention-rate range and for the aggressive-compression regime side by side. \textbf{Bold}: all four deltas non-positive (random weakly outperforms or ties every selector).}
\label{tab:auc-delta-merged}
\end{table*}

\begin{table*}[h]
\centering
\small
\setlength{\tabcolsep}{2.5pt}
\scalebox{0.9}{
\begin{tabular}{ll ccccc ccccc}
\toprule
& & \multicolumn{5}{c}{Full range $r\in[0.01,1.0]$} & \multicolumn{5}{c}{Aggressive compression $r\in[0.01,0.3]$} \\
\cmidrule(lr){3-7}\cmidrule(lr){8-12}
Model & Dataset & rand & $\Delta$low & $\Delta$low\textsubscript{numbers} & $\Delta$numbers & $\Delta$high & random & $\Delta$low & $\Delta$low\textsubscript{numbers} & $\Delta$numbers & $\Delta$high \\
\midrule
Qwen3-4B & AIME24 & 0.52 & \textbf{+0.32} & -0.52 & -0.52 & -0.39 & 0.10 & \textbf{+0.39} & -0.10 & -0.10 & -0.08 \\
 & AIME25 & 0.49 & \textbf{+0.33} & -0.49 & -0.44 & -0.36 & 0.05 & \textbf{+0.38} & -0.05 & -0.02 & -0.05 \\
 & AIME26 & 0.51 & \textbf{+0.36} & -0.51 & -0.51 & -0.40 & 0.05 & \textbf{+0.52} & -0.05 & -0.04 & -0.05 \\
 & MATH-500 & 0.72 & \textbf{+0.13} & -0.33 & -0.40 & -0.18 & 0.46 & \textbf{+0.17} & -0.07 & -0.13 & -0.09 \\
 & GPQA-D & 0.91 & +0.04 & +0.05 & -0.33 & -0.05 & 0.73 & +0.11 & +0.16 & -0.15 & -0.06 \\
 & Zebra & 0.88 & -0.01 & -0.01 & -0.34 & -0.07 & 0.66 & -0.07 & -0.06 & -0.12 & -0.11 \\
\midrule
Qwen3-14B & AIME24 & 0.59 & \textbf{+0.32} & -0.56 & -0.58 & -0.38 & 0.11 & \textbf{+0.57} & -0.09 & -0.10 & -0.09 \\
 & AIME25 & 0.54 & \textbf{+0.32} & -0.54 & -0.54 & -0.34 & 0.07 & \textbf{+0.48} & -0.07 & -0.07 & -0.07 \\
 & AIME26 & 0.55 & \textbf{+0.36} & -0.54 & -0.53 & -0.35 & 0.09 & \textbf{+0.60} & -0.09 & -0.08 & -0.09 \\
 & MATH-500 & 0.88 & \textbf{+0.07} & -0.32 & -0.34 & -0.11 & 0.67 & \textbf{+0.17} & -0.12 & -0.13 & -0.08 \\
 & GPQA-D & 0.91 & \textbf{+0.05} & +0.05 & -0.40 & -0.06 & 0.74 & +0.15 & +0.16 & -0.22 & -0.08 \\
 & Zebra & 0.90 & -0.01 & -0.01 & -0.32 & -0.07 & 0.71 & -0.05 & -0.06 & -0.13 & -0.11 \\
\midrule
gemma-4-E4B-it & AIME24 & 0.35 & \textbf{+0.27} & -0.35 & -0.35 & -0.16 & 0.02 & \textbf{+0.08} & -0.02 & -0.01 & -0.01 \\
 & AIME25 & 0.36 & \textbf{+0.27} & -0.36 & -0.34 & -0.16 & 0.01 & \textbf{+0.08} & -0.01 & -0.00 & -0.01 \\
 & AIME26 & 0.32 & \textbf{+0.31} & -0.32 & -0.32 & -0.10 & 0.00 & \textbf{+0.14} & -0.00 & -0.00 & -0.00 \\
 & MATH-500 & 0.57 & \textbf{+0.12} & -0.29 & -0.31 & -0.07 & 0.29 & \textbf{+0.05} & -0.02 & -0.04 & +0.01 \\
 & GPQA-D & 0.83 & \textbf{+0.01} & -0.09 & -0.41 & -0.02 & 0.58 & \textbf{+0.02} & -0.14 & -0.15 & -0.01 \\
 & Zebra & 0.91 & -0.03 & -0.05 & -0.24 & +0.00 & 0.78 & -0.04 & -0.04 & -0.10 & -0.02 \\
\midrule
gemma-4-26B-A4B-it & AIME24 & 0.53 & -0.04 & -0.04 & +0.01 & -0.05 & 0.14 & -0.12 & -0.14 & -0.14 & -0.08 \\
 & AIME25 & 0.43 & -0.12 & -0.14 & +0.00 & -0.06 & 0.02 & -0.31 & -0.35 & +0.03 & -0.11 \\
 & AIME26 & 0.43 & -0.08 & -0.09 & +0.04 & -0.01 & 0.03 & -0.24 & -0.26 & +0.11 & -0.04 \\
 & MATH-500 & 0.70 & -0.03 & -0.02 & +0.00 & -0.01 & 0.49 & -0.07 & -0.06 & +0.03 & -0.08 \\
 & GPQA-D & 0.94 & -0.01 & -0.15 & -0.22 & -0.00 & 0.83 & -0.02 & -0.25 & -0.10 & -0.01 \\
 & Zebra & 0.93 & -0.02 & -0.12 & -0.33 & -0.01 & 0.78 & -0.05 & -0.21 & -0.19 & -0.01 \\
\midrule
gpt-oss-20b & AIME24 & 0.78 & -0.11 & -0.78 & -0.76 & -0.30 & 0.36 & -0.22 & -0.36 & -0.35 & -0.33 \\
 & AIME25 & 0.72 & -0.07 & -0.72 & -0.72 & -0.28 & 0.29 & -0.17 & -0.29 & -0.29 & -0.28 \\
 & AIME26 & 0.77 & -0.08 & -0.77 & -0.76 & -0.27 & 0.39 & -0.24 & -0.39 & -0.38 & -0.37 \\
 & MATH-500 & 0.81 & -0.04 & -0.53 & -0.53 & -0.22 & 0.51 & -0.09 & -0.23 & -0.24 & -0.22 \\
 & GPQA-D & 0.90 & \textbf{+0.01} & -0.00 & -0.35 & -0.04 & 0.70 & \textbf{+0.03} & +0.01 & -0.15 & -0.09 \\
 & Zebra & 0.83 & \textbf{+0.09} & +0.09 & -0.45 & -0.11 & 0.55 & +0.20 & +0.21 & -0.17 & -0.15 \\
\midrule
gpt-oss-120b & AIME24 & 0.76 & -0.06 & -0.75 & -0.75 & -0.35 & 0.34 & -0.19 & -0.34 & -0.33 & -0.30 \\
 & AIME25 & 0.71 & +0.00 & -0.71 & -0.71 & -0.35 & 0.26 & -0.08 & -0.26 & -0.26 & -0.25 \\
 & AIME26 & 0.73 & +0.00 & -0.73 & -0.72 & -0.33 & 0.27 & -0.07 & -0.27 & -0.26 & -0.25 \\
 & MATH-500 & 0.76 & \textbf{+0.03} & -0.39 & -0.38 & -0.16 & 0.47 & -0.03 & -0.14 & -0.10 & -0.11 \\
 & GPQA-D & 0.87 & -0.00 & -0.00 & -0.43 & -0.06 & 0.63 & -0.01 & +0.00 & -0.21 & -0.07 \\
 & Zebra & 0.85 & \textbf{+0.05} & +0.05 & -0.38 & -0.09 & 0.60 & +0.09 & +0.11 & -0.13 & -0.11 \\
\bottomrule
\end{tabular}
}
\caption{AUC of \emph{random} and delta of each \textbf{token-level} selector relative to it ($\Delta = \mathrm{AUC}_{\text{selector}} - \mathrm{AUC}_{\text{random}}$), \textbf{no patching}, full retention-rate range and aggressive-compression regime. \textbf{Bold} on $\Delta$low: \emph{low-entropy} strictly outperforms both \emph{random} ($\Delta > 0$) and every other selector in that setting.}
\label{tab:token-auc-delta}
\end{table*}

\begin{table*}[t]
\centering
\setlength{\tabcolsep}{2.5pt}
\begin{tabular}{ll cccccc}
\toprule
Model & Dataset & rand & $\Delta$low & $\Delta$low\textsubscript{n} & \textbf{$\Delta$num} & $\Delta$high & $\Delta$newl/$\Delta$eos \\
\midrule
Qwen3-4B & AIME24 & 0.95 & +0.03 & -0.84 & \textbf{+0.05} & -0.43 & -0.89 / -0.92 \\
 & AIME25 & 0.95 & +0.00 & -0.91 & \textbf{+0.04} & -0.52 & -0.95 / -0.95 \\
 & AIME26 & 0.94 & +0.04 & -0.84 & \textbf{+0.05} & -0.41 & -0.94 / -0.94 \\
 & MATH-500 & 0.95 & \textbf{-0.02} & -0.22 & -0.08 & -0.19 & -0.56 / -0.55 \\
 & GPQA-D & 0.98 & \textbf{+0.00} & \textbf{+0.00} & -0.07 & -0.06 & -0.19 / -0.14 \\
 & Zebra & 0.97 & -0.03 & \textbf{-0.02} & -0.41 & -0.09 & -0.39 / -0.29 \\
\midrule
Qwen3-14B & AIME24 & 0.95 & \textbf{+0.04} & -0.64 & \textbf{+0.04} & -0.41 & -0.91 / -0.88 \\
 & AIME25 & 0.95 & +0.03 & -0.82 & \textbf{+0.04} & -0.51 & -0.95 / -0.95 \\
 & AIME26 & 0.95 & +0.03 & -0.68 & \textbf{+0.04} & -0.42 & -0.95 / -0.95 \\
 & MATH-500 & 1.02 & \textbf{-0.03} & -0.14 & -0.05 & -0.11 & -0.44 / -0.43 \\
 & GPQA-D & 0.98 & \textbf{+0.01} & \textbf{+0.01} & -0.14 & -0.06 & -0.15 / -0.12 \\
 & Zebra & 0.97 & \textbf{-0.01} & \textbf{-0.01} & -0.23 & -0.09 & -0.36 / -0.27 \\
\midrule
gemma-4-E4B-it & AIME24 & 0.84 & +0.08 & -0.72 & \textbf{+0.12} & -0.45 & -0.84 / -0.73 \\
 & AIME25 & 0.78 & +0.09 & -0.68 & \textbf{+0.17} & -0.44 & -0.78 / -0.67 \\
 & AIME26 & 0.78 & +0.10 & -0.74 & \textbf{+0.18} & -0.41 & -0.77 / -0.77 \\
 & MATH-500 & 0.87 & \textbf{+0.04} & -0.36 & -0.04 & -0.20 & -0.54 / -0.44 \\
 & GPQA-D & 0.97 & \textbf{-0.00} & -0.04 & -0.08 & -0.08 & -0.11 / -0.25 \\
 & Zebra & 0.96 & \textbf{-0.01} & -0.03 & -0.13 & -0.02 & -0.24 / -0.07 \\
\midrule
gemma-4-26B-A4B-it & AIME24 & 0.96 & +0.05 & -0.28 & \textbf{+0.07} & -0.15 & -0.96 / -0.96 \\
 & AIME25 & 0.93 & +0.02 & -0.54 & \textbf{+0.05} & -0.25 & -0.93 / -0.93 \\
 & AIME26 & 0.94 & +0.03 & -0.39 & \textbf{+0.04} & -0.32 & -0.94 / -0.94 \\
 & MATH-500 & 0.94 & +0.02 & -0.08 & \textbf{+0.03} & -0.07 & -0.38 / -0.48 \\
 & GPQA-D & 0.99 & -0.00 & -0.01 & \textbf{+0.00} & -0.02 & -0.05 / -0.08 \\
 & Zebra & 0.98 & \textbf{+0.01} & +0.00 & -0.12 & -0.03 & -0.06 / -0.18 \\
\midrule
gpt-oss-20b & AIME24 & 0.88 & +0.08 & -0.70 & \textbf{+0.11} & -0.36 & -0.88 / -0.88 \\
 & AIME25 & 0.84 & +0.11 & -0.64 & \textbf{+0.13} & -0.37 & -0.84 / -0.84 \\
 & AIME26 & 0.87 & +0.09 & -0.66 & \textbf{+0.10} & -0.36 & -0.85 / -0.86 \\
 & MATH-500 & 0.91 & \textbf{+0.05} & -0.44 & +0.04 & -0.22 & -0.65 / -0.55 \\
 & GPQA-D & 0.96 & \textbf{+0.01} & -0.01 & -0.06 & -0.04 & -0.30 / +0.01 \\
 & Zebra & 0.91 & \textbf{+0.04} & \textbf{+0.04} & -0.43 & -0.11 & -0.36 / -0.20 \\
\midrule
gpt-oss-120b & AIME24 & 0.87 & +0.08 & -0.28 & \textbf{+0.11} & -0.22 & -0.86 / -0.86 \\
 & AIME25 & 0.85 & +0.10 & -0.18 & \textbf{+0.12} & -0.29 & -0.83 / -0.83 \\
 & AIME26 & 0.86 & +0.08 & -0.15 & \textbf{+0.12} & -0.28 & -0.86 / -0.86 \\
 & MATH-500 & 0.93 & +0.07 & -0.11 & \textbf{+0.10} & -0.17 & -0.62 / -0.49 \\
 & GPQA-D & 0.95 & \textbf{+0.01} & \textbf{+0.01} & -0.11 & -0.01 & -0.15 / +0.00 \\
 & Zebra & 0.93 & +0.03 & \textbf{+0.04} & -0.28 & -0.09 & -0.27 / -0.17 \\
\bottomrule
\end{tabular}
\caption{AUC of \emph{random} and delta of each \textbf{token-level} selector relative to it ($\Delta = \mathrm{AUC}_{\text{selector}} - \mathrm{AUC}_{\text{random}}$), \textbf{patched}, full retention-rate range. Last column: newlines / end-of-sentence deltas. \textbf{Bold} marks the best-performing selector (highest resulting AUC) per row.}
\label{tab:token-auc-delta-patch}
\end{table*}

\end{document}